\documentclass{article}

\usepackage{amsmath, amssymb}
\usepackage{bbm, graphicx, float, pifont,multirow}
\usepackage[printonlyused]{acronym}

\usepackage[squaren,Gray]{SIunits}
\usepackage{pgf,tikz,paralist,mathtools}
\usetikzlibrary{bayesnet,arrows,automata}
\usepackage[normalem]{ulem}

\usepackage{setspace}
\usepackage[left=3.5cm,right=3.5cm,top=3cm,bottom=4cm]{geometry}

\newcommand{\cf}{c.f.~}
\newcommand{\eg}{e.g.,~}
\newcommand{\ie}{i.e.,~}
\newcommand{\ia}{i.a.,~}

\newcommand{\SNR}{\text{SNR}}

\newcommand{\dlong}{\text{long}}
\newcommand{\dlat}{\text{lat}}

\newcommand{\iid}{i.i.d.~}

\newcommand{\rreal}{\mathbb{R}}

\newcommand{\rrealplusnull}{\rreal_+}

\newcommand{\diag}[1]{\text{diag}{\left(#1\right)\,}}

\newcommand{\vs}[1]{\boldsymbol{\mathrm{#1}}}

\newcommand{\eeqref}[1]{Eq.~(\ref{#1})}
\newcommand{\figref}[1]{Fig.~\ref{#1}}

\newcommand{\secref}[1]{Section~\ref{#1}}

\newcommand{\tabref}[1]{Tab.~\ref{#1}}

\newcommand{\expf}[1]{\exp\{#1\}}

\newcommand{\numObs}{ {N_\text{z}} }

\newcommand{\stdObs}{\sigma_\text{z}}

\newcommand{\proppdf}[1]{q(#1)}

\newcommand{\Xmat}[0]{\vs X}

\newcommand{\obsDim}{D}

\newcommand{\numAct}{ {N_\text{u}} }

\newcommand{\hyperPhiAlpha}[1]{ {\alpha_{\phi#1}} }
\newcommand{\timeStamp}[1]{ t_{\text{F}#1}}
\newcommand{\sampleFreq}{ {f_\text{s}} }

\newcommand{\act}{u}
\newcommand{\actv}{\vs u}

\newcommand{\hyperSthetav}{ {\vs{\theta}_s} }
\newcommand{\hyperSthetaZero}{ {\theta_{s=0}} }
\newcommand{\hyperSthetaNonZero}{ {\theta_{s\neq0}} }

\newcommand{\accuracyAct}{A_\text{u}}

\newcommand{\Actset}{ \mathcal{U} }

\newcommand{\onevec}[1]{\vs 1_#1}

\newcommand{\prodNobs}[0]{\prod_{n=1}^{\numObs}}

\newcommand{\prodD}[0]{\prod_{d=1}^D}
\newcommand{\prodK}[0]{\prod_{k=1}^K}

\newcommand{\sumact}[0]{\sum_{\act \in \Actset}}

\newcommand{\sumNobs}[0]{\sum_{n=1}^\numObs}
\newcommand{\sumD}[0]{\sum_{d=1}^D}

\newcommand{\sumK}[0]{\sum_{k=1}^K}

\newcommand{\sumKtwok}[0]{\sum_{\substack{k'=1\\k' \neq k}}^K}

\newcommand{\pdf}[0]{p}
\newcommand{\pms}[0]{P}

\newcommand{\intd}{\mathrm{d}}

\newcommand{\betabinomialpdf}[2]{ \text{BetaBin}_{#1}\!\left(#2\right)\,}

\newcommand{\catpdf}[2]{ \text{Cat}_{#1}\!\left(#2\right)\,}
\newcommand{\dirpdf}[2]{ \text{Dir}_{#1}\!\left(#2\right)\,}
\newcommand{\dirmultpdf}[2]{ \text{DirMult}_{#1}\!\left(#2\right)\,}
\newcommand{\exppdf}[2]{ \text{Exp}_{#1}\!\left(#2\right)\,}
\newcommand{\gampdf}[2]{ \text{Ga}_{#1}\!\left(#2\right)\,}
\newcommand{\gausspdf}[3]{\mathcal{N}_{#1}\!\left(#2,#3\right)\,}
\newcommand{\invgampdf}[2]{ \text{IGa}_{#1}\!\left(#2\right)\,}

\newcommand{\poissonpdf}[2]{ \text{Poisson}_{#1}\!\left(#2\right)\,}
\newcommand{\truncgausspdf}[2]{\mathcal{TN}_{#1}\!\left(#2\right)\,}

\newcommand{\transSym}{\text{T}}
\newcommand{\trans}[1]{{#1}^{\transSym}}

\newcommand{\betafun}[1]{\text{B}\!\left(#1\right)\,}

\newcommand{\kroneckerfun}[1]{\delta\!\left(#1\right)\,}

\newcommand{\Amat}{ {\vs A} }
\newcommand{\AFmat}{ {\vs A^\star} }

\newcommand{\Imat}{ {\vs I} }
\newcommand{\Fmat}{ {\vs F} }

\newcommand{\Smat}{ {\vs S} }

\newcommand{\Wmat}{ {\vs W} } 
\newcommand{\Zmat}{ {\vs Z} }

\newcommand{\Phimat}{ {\vs \Phi} }

\newcommand{\MAP}{ \text{MAP} }

\newcommand{\gtXmat}{ \vs{\check{X}}}
\newcommand{\gtFmat}{ \vs{\check{F}}}
\newcommand{\gtWmat}{ \vs{\check{W}}}

\newcommand{\gthyperWgamma}{ \check{\gamma}_\text{w} }
\newcommand{\gtPhimat}{\check{\vs{ \Phi}}}

\newcommand{\gtSmat}{\check{\vs S}}

\newcommand{\gtSigmaBNFL}{{\check{\sigma}_\text{z}}}

\newcommand{\sigmaZ}{ \sigma_{\text{z}} }

\newcommand{\hyperSalphaA}{ {\alpha_s} }
\newcommand{\hyperSalphaB}{ {\beta_s} }

\newcommand{\hyphypPhiA}{ {h^{(1)}_{\phi_A}} } 
\newcommand{\hyphypPhiB}{ {h^{(2)}_{\phi_A}} }

\newcommand{\hyphypAalphaA}{ {h^{(1)}_{\alpha_A}} } 
\newcommand{\hyphypAalphaB}{ {h^{(2)}_{\alpha_A}} }
\newcommand{\hyphypAbetaA}{ {h^{(1)}_{\beta_A}} } 
\newcommand{\hyphypAbetaB}{ {h^{(2)}_{\beta_A}} }

\newcommand{\hyphypSigmaAlphaA}{ {h^{(1)}_{\alpha_\sigma}} }
\newcommand{\hyphypSigmaAlphaB}{ {h^{(2)}_{\alpha_\sigma}} }
\newcommand{\hyphypSigmaBetaA}{ {h^{(1)}_{\beta_\sigma}} }
\newcommand{\hyphypSigmaBetaB}{ {h^{(2)}_{\beta_\sigma}} }

\newcommand{\gthyphypWgammaA}{ {\check{h}_{\alpha_\gamma}} }
\newcommand{\gthyphypWgammaB}{ {\check{h}_{\beta_\gamma}} }
\newcommand{\hyphypWgammaA}{ {\alpha_\gamma} }
\newcommand{\hyphypWgammaB}{ {\beta_\gamma} }

\newcommand{\hyperAalpha}{ {\alpha_a} }
\newcommand{\hyperAbeta}{ {\beta_a} }

\newcommand{\hypertauA}{ {\alpha_\sigma} }
\newcommand{\hypertauB}{ {\beta_\sigma} }

\newcommand{\hyperWgamma}{ {\gamma_\text{w}} }

\newcommand{\Dset}{ \mathcal{D} }
\newcommand{\Fset}{ \mathcal{F} } 
\newcommand{\Sset}{ \mathcal{S} }
\newcommand{\Xset}{ \mathcal{X} } 
\newcommand{\Zset}{ \mathcal{Z} }

\newcommand{\av}[0]{\vs a}
\newcommand{\fv}[0]{\vs f}
\newcommand{\sv}[0]{\vs s}
\newcommand{\tv}[0]{\vs t}
\newcommand{\wv}[0]{\vs w}
\newcommand{\xv}[0]{\vs x}

\newcommand{\zv}[0]{\vs z}

\newcommand{\phiv}[0]{\vs \phi}
\newcommand{\thetav}[0]{\vs \theta}

 \newcommand{\abs}[1]{\left|\,#1\,\right|}

\acrodef{CFA}{Contingent Feature Analysis}
\acrodef{CRP}{Chinese Restaurant Process}
\acrodef{DL}{Deep Learning}
\acrodef{FMDP}{Factored Markov Decision Process}
\acrodef{IBP}{Indian Buffet Process}
\acrodef{IMU}{Inertial Measurement Unit}	
\acrodef{IRL}{Inverse Reinforcement Learning}
\acrodef{LFD}{Learning From Demonstrations}
\acrodef{LIDAR}{Light Detection And Ranging}
\acrodef{MAP}{\textit{maximum-a-posteriori}}
\acrodef{MAD}{Mean Absolute Deviation}
\acrodef{MDP}{Markov Decision Process}
\acrodef{MMSE}{Minimum Mean Squared Error}
\acrodef{MSE}{Mean Squared Error}
\acrodef{NMF}{Non-negative Matrix Factorization}
\acrodef{PCA}{Principal Component Analysis}
\acrodef{pdf}{probability density function}
\acrodef{pmf}{probability mass function}
\acrodef{POMDP}{Partially Observable Markov Decision Process}
\acrodef{RL}{Reinforcement Learning}
\acrodef{RMSE}{Root Mean Squared Error}
\acrodef{SNR}{Signal-To-Noise Ratio}

\pdfoutput=1

\begin{document}

\title{Bayesian Nonparametric Feature and Policy Learning for Decision-Making}

\author{J{\"u}rgen~Hahn\footnote{jhahn@spg.tu-darmstadt.de} \quad 
	Abdelhak~M.~Zoubir\footnote{zoubir@spg.tu-darmstadt.de}\\
    Signal Processing Group\\ Institute of Telecommunications, Technische Universit{\"a}t Darmstadt\\ Merckstra{\ss}e 25, 64283 Darmstadt}
\date{}

\maketitle

\begin{abstract}
Learning from demonstrations has gained increasing interest in the recent past, enabling an agent to learn how to make
decisions by observing an experienced teacher. 
While many approaches have been proposed to solve this problem, there is only little work that focuses on reasoning about the observed
behavior.  
We assume that, in many practical problems, an agent makes its decision based on latent features, indicating a certain
action.
Therefore, we propose a generative model for the states and actions. 
Inference reveals the number of features, the features, and the policies, allowing us to learn and to analyze the
underlying structure of the observed behavior.
Further, our approach enables prediction of actions for new states.
Simulations are used to assess the performance of the algorithm based upon this model. 
Moreover, the problem of learning a driver's behavior is investigated, demonstrating the performance of the proposed model in a real-world scenario.
\end{abstract}

\section{Introduction}
Decision-making plays a crucial role in many applications, such as robot learning, driver assistance systems, and recommender systems \cite{Shani2002}. 
A fundamental question in decision-making is how an agent can learn to make optimal decisions.
Learning from an experienced teacher provides a natural means to solve this problem, without the need of explicitly defining rules for the desired behavior.
Further, observing a teacher may provide a deeper understanding of the decision-making process. 
Therefore, \ac{LFD} \cite{Argall2009} has gained a lot of interest in the recent past.

According to \cite{Argall2009}, approaches for \ac{LFD} can be grouped into (i) reward-based models and (ii) imitation learning.
In reward-based models, it is assumed that the agent makes its decision based on a reward which is, in the context of \ac{LFD}, learned from observations (as in \ac{IRL} \cite{Ng2000}).
In imitation learning, it is assumed that an experienced teacher can be observed.
Thus, the policy, telling the agent how to act in a given situation, can be learned by understanding the direct relation between the teacher's states and actions.
Especially in reward-based approaches, problems defined on infinite spaces, where the state of an agent can take continuous values, are mostly intractable to solve and efficient approximations are needed.
Only in the case of finite state and action spaces and known rewards, learning optimal policies has been solved \cite{Bellman1957}.
A typical approach to facilitate this problem is the representation of the state space in a feature domain, \cf \cite{Pomerleau1991,Bradtke1996}. 

However, decision-making has been viewed only from a limited feature-based perspective, where features are usually designed by experts and mainly serve the purpose of reducing the size of the state space.
We argue that, in many practical systems, the agent makes its decision based on a compact representation of the observed data, which can be considered as a projection onto a feature space of the decision-making problem.
As an example, consider a person driving a car. 
The driver's observations consists of, \ia the location, speed, and acceleration of his vehicle and the surrounding vehicles, the type of the road and the weather conditions.
However, the driver makes his decision based on a subset of the available information, \eg on the time-to-reach between his and the other road users' cars.
This idea has also been investigated from a psychological point of view by the concept of discovering latent causes in human behavior, which is related to learning state space representations \cite{Gershman2015}.

Assuming that a certain structure underlies the observations, we aim at inferring the latent features and build a feature-based representation of the states, yielding the following two advantages:
first, the states can be represented compactly, rendering the decision-making problem much more efficient as compared to working in the original domain.
Second, the features can be regarded as causes for the observed decisions, allowing for a deeper understanding of the observed behavior. 
In particular, we consider a \ac{LFD} problem and propose a Bayesian nonparametric framework for feature learning in \ac{LFD}. 
We assume that the policies depend on the features of the observed demonstrations. 
With this model, we are able to (i) significantly reduce the state space by (ii) learning the features as well as the number of features and (iii) provide a better understanding of what caused the teacher to take the observed actions.

In the next parts of this section, we formulate the problem and review the state-of-the-art for feature learning in \ac{LFD}. 
In \secref{sec::model}, we motivate and present the model for feature-based decision-making. 
The model is detailed in \secref{sec::dm::model} and we explain how the number of latent features can be inferred from the data. 
\secref{sec::inference} presents the inference scheme for the latent variables. 
The proposed algorithm is empirically evaluated with simulation experiments in \secref{sec::simulations}, demonstrating its performance.
Real data experiments are conducted in \secref{sec::realdata}.
We discuss our findings in \secref{sec::discussion} and end with a short conclusion in \secref{sec::conclusion}.

\subsection{Problem formulation}
The goal of this work is to introduce a feature-based \ac{LFD} framework. 
While this model can be used for teaching an agent given observations of the desired behavior, the focus of this work lies on the analysis of the observed behavior by means of the features.
Since we consider a decision-making task, the investigated problem can be modeled by means of a \ac{POMDP} \cite{Kaelbling1998,Smallwood1973}, which is defined by
\begin{itemize}  
  \item a set of observations, $\Zset$,
  \item a set of states, $\Xset$,
  \item a finite set of $\numAct$ actions, $\Actset$, 
  \item a transition model, which describes the probability of entering a state after taking an action in the current state,
  \item an observation model which explains how the observations are generated from the states,
  \item a discount factor, which penalizes long-term rewards,
  \item and a reward function, $R$. 
\end{itemize}
As the main goal of this work is to provide a means to understand observed behavior, we consider an imitation learning approach and learn the relation between states and actions from the observations directly, where the reward and, hence, the discount factor, are not considered. 

We assume that we (or an agent) have access to $\numObs$ noisy observations, $\zv_n \in \Zset$, of the states, $\xv_n \in \Xset$, with $n=1,\ldots,\numObs$. In particular, we consider Gaussian noise, \ie $\zv_n$, describes observations of the states, $\xv_n$, with additive Gaussian noise. Further, we assume that the actions, $\act_n \in \Actset$,  taken in the corresponding states, can be observed.
The observations are assumed to be optimal in the sense that they represent the behavior of the agent seeking its goal, \ie without any exploratory steps.

A simple approach to the considered problem would be the use of a feature extraction technique such as \ac{PCA} \cite{Jolliffe2002} or \ac{NMF} \cite{Lee2001} to learn features from the observed states. 
Following this solution, features cannot be jointly learned and shared between different actions. Moreover, learning discriminative features is not guaranteed. 
Therefore, we argue that the features and policies need to be learned jointly such that a trade-off between feature sharing, promoting a compact model, and discrimination capability is found based on the observations.

\subsection{Relation to existing work}
In the following, we provide an overview about the current state of the art of feature learning for \ac{LFD}. As \ac{IRL} usually requires to solve an \ac{MDP}, which is usually done by means of a \ac{RL} algorithm, we start with an introduction in feature learning for \ac{RL}.

\subsubsection{Reinforcement Learning}
Large state and action spaces are especially problematic for value-based \ac{RL} algorithms such as value-iteration \cite{Bellman1957} or Q-learning \cite{Watkins1989} since the value function, representing the expected accumulated reward, needs to be approximated.
In early approaches, a set of basis functions, often referred to as features, is linearly weighted to represent the this function \cite{Bradtke1996}. 
However, there exists only little work on learning these basis functions. 
Riedmiller \cite{Riedmiller2005} has proposed the Q-fitted value iteration where the value function is approximated by means of a neural network, where the features are learned in the layers of the network.
In \cite{Mnih2015}, this approach has been extended by replacing the neural network by a deep-layered counterpart. 
A different concept to employ features is proposed by Hutter with the Feature Reinforcement Learning framework \cite{Hutter2009}. 
In this framework, the goal is to learn a feature mapping from the agent's history (comprised of actions, states, and rewards) to a \ac{MDP} state, enabling decision learning for infinite state spaces \cite{Daswani2014}. 
An alternative framework for learning the latent structure of the state space is proposed in \cite{Degris2006} which is based on \acp{FMDP} \cite{Boutilier1995}. 
In the \ac{FMDP} framework, it is assumed that the observed states can be represented compactly by exploiting the structure within the states, enabling efficient learning.
This approach has been extended in \cite{Nguyen2013} to an online approach, where the features are selected from a large set by means of Group LASSO.
In these approaches, the inferred features mainly serve the purpose of dimensionality reduction.
Therefore, the features do not necessarily possess a meaning that can be easily interpreted. 

\subsubsection{Inverse Reinforcement Learning}
\ac{IRL} is concerned with the problem of learning the reward function from observed behavior \cite{Ng2000}. 
As in \ac{RL}, features are often used to linearly parameterize the reward function, \eg in \cite{Ng2000, Hahn2015}. 
Recent attempts have been made to consider a \ac{DL} architecture \cite{Wulfmeier2015} for \ac{IRL}, providing means for nonlinear, hierarchical feature learning.

A Bayesian nonparametric approach is proposed in \cite{Choi2012}, utilizing an \ac{IBP} to model feature activations. 
As the features of the reward function are assumed to be known, this approach can be understood rather as a feature selection than feature learning for \ac{IRL}, where the number of features is inferred by the \ac{IBP}. 
Different results on Bayesian nonparametrics for \ac{IRL}, which is indirectly related to feature learning, are given in \cite{Michini2012}, where a partitioning of the state space is sought for, or \cite{Surana2014}, where complex behavior is decomposed into several, simpler behaviors that can be easily learned.

\subsubsection{Imitation Learning}
Instead of estimating the reward as in \ac{IRL}, imitation learning aims at inferring the underlying policy directly \cite{Atkeson1997, Sosic2016a}.
Usually, handcrafted features are used, \eg in \cite{Ratliff2006, Ross2011}. 
Attempts to introduce new features are made in \cite{Ratliff2007} as an extension of the maximum margin planning algorithm which is proposed in \cite{Ratliff2006}.
As explained in \cite{Argall2009}, imitation learning can be considered as a supervised learning task. 
Thus, feature selection and learning techniques developed for classification and regression can also be used in imitation learning. 
An excellent overview is given in \cite{Guyon2003}.
Although these models work well in practice, they might not be able to provide a deeper understanding of the observed behavior as they do not explicitly model the states and actions.

\section{Choice of the Model}
\label{sec::model}
In the first part of this section, we propose a feature model for \ac{LFD}. In the second part, we explain the relevance of the transition model to the proposed framework. Since we assume a mixture of policies in this framework, we briefly discuss the intuition behind this assumption in the third part. Alternative models for feature learning for \ac{LFD} are discussed in the forth part.

\subsection{Feature model for learning from demonstrations}
\label{sec::model::fl}
We assume a linear latent feature model, similar to \ac{NMF} \cite{Lee2001} and \ac{PCA} \cite{Jolliffe2002}. 
Thus, the noisy observations, $\zv_n \in \rreal^{1 \times D}, n=1,\ldots,\numObs,$ are assumed to be composed of the latent features, $\Fmat \in \Fset^{K \times D}$, and the feature coefficients, $\sv_n \in \Sset^{1 \times K}$,
\begin{align}
  \zv_n = \sv_n \Fmat + \vs \epsilon_n,
  \label{eq::choice::observation}
\end{align}
where $\vs \epsilon_n$ represents Gaussian \iid noise with variance $\stdObs^2$ and the states are given as $\xv_n =  \sv_n \Fmat$. The number of features is $K$ and the dimension of the observations is $D$.
Clearly, the feature space, $\Fset$, depends on the application.
In the following, we assume that the features are positive-valued, \ie $\Fset = \rrealplusnull$.
Following \cite{Knowles2011}, the feature matrix is composed of a binary activation matrix, $\Amat \in \{0, 1\}^{K \times D}$, and a weighting matrix, $\Wmat \in \Fset^{K \times D},$ where the relation is given by the Hadamard product,
\begin{align}
  \Fmat = \Amat \odot \Wmat.
  \label{eq::dm::feature::model}
\end{align}
The feature model in \eeqref{eq::dm::feature::model} can be easily extended to an infinite feature model by placing an \ac{IBP} \cite{Ghahramani2005,Griffiths2011} prior over $\Amat$. 
The \ac{IBP} assumes an infinite number of features, while the observed data can be explained by a finite number. 
This gives rise to a nonparametric model, where the number of features is implicitly modeled by means of the \ac{IBP}. This is detailed in \secref{sec::dm::model}.

A fundamental difference to most existing work on decision-making using feature representations is that we assume that the agent makes its decision based on features, where each feature attracts the agent to take a specific action.
We consider a (latent) linear feature model where the feature coefficients depend on the observed state and determine the actions. 
For this reason, we refer to the feature coefficients also as \emph{substates}.

\subsection{Transition model}
In decision-making problems, the transition model explains which actions the agent can take in each state.
Thus, the transition model acts as a constraint on the possible actions. 
As we focus on inferring the latent causes for the observed actions by means of features, the transition model is less relevant as it only provides additional information.
Besides, the transition model is rarely known in practice.
As we are given observations, we can, theoretically, infer the transition model. For this, we could either employ a parametric or a nonparametric model. 
Defining a parametric model for the transitions is not trivial, as the dynamics in the latent space may be highly nonlinear.
Alternatively, a nonparametric model can be assumed. This, however, requires a large amount of observations for the estimation of the parameters, which is often not available.
Assuming that the noise in the observations is low, we argue that we can reliably infer the substates from the corresponding observations, eliminating the need for a transition model.

\subsection{Feature-based policy}
\label{sec::intro::policy}
As each feature imposes its on own policy, the probability of the agent taking an action, $\act$, in a substate, $\sv$, is a mixture of the feature policies, $\pms( \act \,|\, \phiv_k)$,
\begin{align}
   \pms( \act \,|\, \sv, \Phimat) \propto \sumK s_k \pms( \act \,|\, \phiv_k), 
   \label{eq::dm::policy::model}
\end{align} 
where $\phiv_k$, $k=1,\ldots,K$, are the parameters of the feature policies and $\Phimat = [\phiv_1,\ldots,\phiv_K]$.
The mixture of policies  can be interpreted either as a stochastic or a deterministic policy. 
In the first case, the action to be taken should be sampled according to \eeqref{eq::dm::policy::model}.
In the second case, simply the most probable action is taken.
Mixture polices have been investigated in multi-objective problems, where an agent aims at reaching several objectives, some of which can even be conflicting \cite{Parisi2014,Vamplew2009,Shelton2001}. 
A stochastic policy is needed in this framework to ensure that all goals can be satisfied.
Similar problems occur if the problem at hand is described by a \ac{POMDP}, where the true state of the agent is unknown. 
The uncertainty about the state of the agent is expressed by beliefs over states. 
Acting according to a stochastic policy maximizes the expected return \cite{Kaelbling1998}.  

In case of single agents, it has been shown that deterministic policies are optimal solutions for \ac{MDP}s \cite{Puterman1994}.
As explained in \secref{sec::model::fl}, we argue that we can reliably infer the substates.
Thus, we assume a deterministic policy in the following, where the probability in \eeqref{eq::dm::policy::model} expresses our confidence about the actions given the states. 

Note that we also assume that each feature imposes a deterministic policy such that we expect a strongly peaked distribution for the feature policies, expressing the confidence of the chosen action.

\subsection{Alternative Feature-based Models}
Of course, there are other possibilities to incorporate feature learning in \ac{LFD}. We briefly discuss two alternatives with their potential advantages and drawbacks. For completeness, as mentioned in the introduction, in the most trivial setup, one could simply cluster the states according to the observed actions and then learn the features. This, however, has the significant drawback that the features cannot be shared between the clusters.

\subsubsection{Unique coefficient model}
Instead of assuming, as in our model, that the features determine the behavior of the agent, the feature coefficients can be clustered, where the clusters indicate the optimal actions given the coefficients. 
Thus, the clusters can be interpreted as latent substates.
Supposing that the elements of the feature coefficients are binary-valued, we can convert each feature coefficient vector to a unique identifier, representing the cluster.
Thus, instead of employing expensive clustering, a fast deterministic mapping from the binary coefficients to the cluster identifier can be used.
However, as the identifiers must be unique for the cluster assignments, we can have at maximum $2^K$ different clusters in this setup, \ie the number of clusters (and, hence, possible actions) is strictly limited by the number of features.
An advantage of this model is that the relation between substates and policies can be nonlinear.
However, a significant drawback is that this model suffers severely from errors in the inference of the substates. 
Consider the case, where, for instance, due to noise, one element of the substate is incorrectly set.
This substate is then assigned to a new cluster, for which the policy must be inferred from potentially little data, yielding highly varying policy estimates.
As in our approach, an \ac{IBP} prior can be placed over the feature activations to infer the number of latent features.

\subsubsection{Clustering-based approach}
A clustering-based approach assumes that similar states can be grouped and result in the same behavior. 
This can also be understood as a single feature model, in which we assume that the observations can be described by a single feature, representing the clusters.
Thus, the substates reduce to cluster indicators, \ie they indicate which feature best represents the observed state.
The number of latent states is then equal to the number of features. 
One possibility to infer this number is to utilize a \ac{CRP}, giving rise to a Bayesian nonparametric model \cite{Michini2012}. 
A similar model, where the state space is clustered according to the played actions, is proposed in \cite{Sosic2016}. 

\subsubsection{Relation to the proposed model}
Our model has the advantage, as opposed to the clustering-based approach and the unique coefficient model, that the features provide a means to understand the observed behavior.
In contrast to the unique coefficient model, we neither require binary coefficients nor a clustering step.
Compared to the clustering-based approach, our model is able to significantly reduce the number of latent features and policies, as the features can be shared by different states.
However, our model suffers from the assumption of a linear relation between features, substates, and policies. 
This problem is alleviated in the other approaches, as the features are decoupled from the policies.
Note that our model becomes similar to the clustering-based model at the price of higher computational costs, if each substate is represented by only one feature.

\section{Bayesian Nonparametric Model for Feature Learning}
\label{sec::dm::model}
In this section, we provide a general framework for Bayesian nonparametric feature learning for decision-making based on the model proposed in \secref{sec::model::fl}. In order to learn the structure, we assume that we are given a set of observations consisting of state-action pairs, $\Dset =\{ (\zv_1, \act_1), \ldots, (\zv_\numObs, \act_\numObs) \} $. 

\subsection{Observation likelihood and noise variance}
The observations, $\zv_n, n=1,\ldots,\numObs$, are assumed to be conditionally independent. 
As we assume Gaussian noise in \eeqref{eq::choice::observation}, the state likelihood can be expressed as
\begin{align}
 \pdf(\Zmat \, | \, \Wmat, \Amat, \Smat, \sigmaZ^2) = \prodNobs \gausspdf{\zv_n}{\sv_n \left( \Amat \odot \Wmat\right)}{\sigmaZ^2 \Imat}.
 \label{eq::bnfl::likelihood}
\end{align}
with $\Zmat = \trans{\begin{bmatrix} \trans{\zv_1} & \ldots & \trans{\zv_\numObs} \end{bmatrix}}$ and $\Smat = \trans{\begin{bmatrix} \trans{\sv_1} & \ldots & \trans{\sv_\numObs} \end{bmatrix}}$.
The variance of the noise, $\sigmaZ^2$, is assumed to be Inverse-Gamma distributed with hyperparameters $\hypertauA, \hypertauB$.
Further, we place hyperpriors on $\hypertauA$ and $\hypertauB$, following Gamma distributions with hyperparameters $\hyphypSigmaAlphaA, \hyphypSigmaAlphaB, \hyphypSigmaBetaA$, and $\hyphypSigmaBetaB$.

\subsection{Prior for the feature weights}
As explained above, the prior probability of the feature weights, $\Wmat$, depends on the problem at hand.
We consider \iid positive-valued feature weights, $w_{k,d}$ with $k=1,\ldots,K$ and $d=1,\ldots,D$, and assume an Exponential prior, 
\begin{align*}
 \pdf(\Wmat \,|\, \hyperWgamma) = \prodK \prodD \exppdf{w_{k,d}}{\hyperWgamma}.
\end{align*}
The scaling factor, $\hyperWgamma$, is assumed to be Inverse-Gamma distributed with hyperparameters $\hyphypWgammaA$ and $\hyphypWgammaB$. 

If the features are assumed to be real-valued, the prior can be modeled with a Gaussian distribution with straightforward modifications.

\subsection{Prior for the feature activations}
The feature activations are modeled by means of an \ac{IBP} \cite{Ghahramani2005,Griffiths2011}, assuming an infinite number of features. 
In the following, we consider the two-parameter generalization \cite{Ghahramani2007} which allows to sample sparse as well as dense matrices.

The \ac{IBP} is derived as follows.
For a finite number of features, $K^\star$, the sums over the rows of the feature activation matrix, $\AFmat \in \{0,1\}^{K^\star \times D}$, are assumed to follow \iid Binomial distributions.
Placing a Beta prior with hyperparameters $\frac{\hyperAalpha \hyperAbeta}{K^\star}$ and $\hyperAbeta$ over the parameter of the Binomial distribution and marginalizing over this parameter yields a Beta-Binomial distribution \cite{Ghahramani2005, Ghahramani2007}.
Since we are interested in sampling from an infinite number of features, we consider the limit for $K^\star \rightarrow \infty$, resulting in the distribution of the activation matrix $\Amat$ \cite{Ghahramani2005,Ghahramani2007}, 
\begin{align}
 \begin{split}
 \pms(\Amat\,|\,\hyperAalpha,\hyperAbeta) &\coloneqq \lim_{K^\star \rightarrow \infty} \pms(\AFmat\,|\,\hyperAalpha,\hyperAbeta)
 \\ & = \frac{(\hyperAalpha \hyperAbeta)^{K}}{\prod_{\vs h \in \{0,1\}^D \backslash \vs 0} K_h!} \expf{-\bar{K}}
   \prodK \betafun{m_k, D - m_k + \hyperAbeta}, 
 \end{split}
 \label{eq::ibp::infinitemodel}
\end{align}
where the number of occurrences of the binary vector $\vs h \in \{0,1\}^D$ is denoted by $K_h$ and $\text{B}$ is the Beta function.
Note that $\Amat$ is a matrix with infinitely many rows. However, due to the sparsity assumption, only a finite number of the rows of the realizations contain active elements, denoted by $K$. Thus, we need to store only rows with active elements in memory, which can be understood as realizations of $K$ features, where
the average number of active rows is given by $\bar{K} = \hyperAalpha \sumD \frac{\hyperAbeta}{\hyperAbeta + d -1}$ \cite{Ghahramani2005, Doshi-Velez2009}. 
The hyperparameters $\hyperAalpha$ and $\hyperAbeta$ reflect our prior belief about the number of features and the sparsity of the matrix \cite{Ghahramani2007}.
The probability of activating an element increases with both hyperparameters.
As the number of expected features grows linearly with $\hyperAalpha$, $\hyperAalpha$ can be used to control the number of generated features. 
Considering the limits for $\hyperAbeta \rightarrow 0$ and $\hyperAbeta \rightarrow \infty$ shows that the expected number of features is limited to $\hyperAalpha D$. As the probability of an element of $\Amat$ being active increases, $\hyperAbeta$ can be understood as a means to control the sparsity of the realizations. 
We infer $\hyperAalpha$ and $\hyperAbeta$ from the observations, assuming that they follow Gamma distributions, \ie \mbox{$\hyperAalpha \sim \gampdf{\hyperAalpha}{\hyphypAalphaA, \hyphypAalphaB}$} and \mbox{$\hyperAbeta \sim \gampdf{\hyperAbeta}{\hyphypAbetaA, \hyphypAbetaB}$} \cite{Knowles2011}.

\subsection{Prior for the substates}
\label{sec::model::smat}
As formulated in \eeqref{eq::choice::observation}, we assume that the observations are composed of a mixture of features weighted by the substates. 
Similar to a \ac{FMDP}\footnote{A fundamental difference between our model and a \ac{FMDP} is that the substates in our model are, theoretically, not restricted to be finite as we do not need to enumerate over them.}, we restrict the domain of the substate elements, $s_{n,k}$, with $n=1,\ldots,\numObs$ and $k=1,\ldots,K$, to take values from a finite set, $\Sset = \{ \breve{s}_1, \ldots \breve{s}_L\}$, where $L$ denotes the number of elements, to simplify inference. For convenience, we assume equidistant elements in $\Sset$ and set $\breve{s}_1 = 0$ and $\breve{s}_L = 1$. Note that the limited range does not restrict the model, as the features can be scaled to fit the observations.

Further, we assume that the substates are sparse, meaning that each observation consists of only a few features such that only few polices determine the observed action. 
In particular, we consider a sparsity-promoting mixture prior on the substates, similar to a \emph{spike} and \emph{slab} model \cite{Mitchell1988,Ishwaran2005}. 
The components are given by a Categorical distribution, where $P(s = 0\,|\,\theta_{s=0})$ is the sparsity component and $P(s \neq 0\,|\,\theta_{s\neq0})$ is the weight component. Note that all categories, except $\breve{s}_1 = 0$, have equal probability. 
We place a Beta prior, $\pdf(\hyperSthetav \,|\, \frac{\hyperSalphaA_{=0}}{2}, \frac{\hyperSalphaA_{\neq0}}{2})$, with hyperparameters $\hyperSalphaA_{=0}$ and $\hyperSalphaA_{\neq0}$, over the mixture weights $ \hyperSthetav = \{\hyperSthetaZero, \hyperSthetaNonZero\}$ with $\hyperSthetaNonZero = 1-\hyperSthetaZero$. Marginalizing over $\hyperSthetaZero$ yields,
\begin{align*}
    \pms(\Smat) & = \prodK \int_0^1 \prodNobs \bigg( \pms(s_{n,k}=0\,|\,\hyperSthetaZero) \kroneckerfun{s_{n,k}} 
 \\             & \hspace{2cm}  + \pms(s_{n,k}\neq0\,|\,\hyperSthetaZero) (1- \kroneckerfun{s_{n,k}}) \bigg) 
 \\             & \hspace{2cm}  \times \pdf(\thetav \,|\, \hyperSalphaA_{=0},\hyperSalphaA_{\neq0}) \, \intd\theta_{s=0}
 \\             & \propto  \prodK \betabinomialpdf{\sv_k}{ m_{s=0,k} + \hyperSalphaA_{=0}, m_{s\neq0,k} + \hyperSalphaA_{\neq0}},
\end{align*}
where $\kroneckerfun{s}$ returns one if $s=0$ and zero otherwise, $m_{s=0,k}$ counts the zero elements in $\sv_k$, and $m_{s\neq0,k} = \numObs - m_{s=0,k}$. 

\subsection{Mixture of policies and action likelihood}
Since we assume a finite set of actions, we consider a Categorical distribution for each mixture component, 
\begin{align*}
 \pms(\act\,|\,\phiv_k) = \catpdf{\act}{\phiv_k},
\end{align*}
such that the mixture model can be written as
\begin{align}
 \pms(\act \,|\,\Phimat,\sv) = \frac{1}{Z_\act} \sumK s_k \pms(\act\,|\,\phiv_k), 
 \label{eq::model::mixture}
\end{align}
with normalization constant $Z_\act = \sumact \sumK s_k \pms(\act\,|\,\phiv_k)$.
The parameters of the policy of each mixture component follow Dirichlet distributions with identical hyperparameters, $\hyperPhiAlpha{}$,
\begin{align*}
 \pdf(\Phimat) = \prodK \dirpdf{\phiv_k}{\hyperPhiAlpha{},\ldots,\hyperPhiAlpha{}},
\end{align*}
where we assume independent policies. 
We consider the hyperparameter, $\hyperPhiAlpha{}$, as a Gamma distributed variable with parameters $\hyphypPhiA{}$ and $\hyphypPhiB$. 

As explained, we require a data set containing observed actions, $\act_n$, $n=1,\ldots,\numObs$.
Assuming that the observed actions are independent, the likelihood is given as
\begin{align*}
 \pms(\actv \,|\,\Phimat,\Smat) =  \prodNobs \frac{1}{Z_{\act_n}} \sumK s_k \pms(\act_n\,|\,\phiv_k).
\end{align*}

\subsection{Joint posterior distribution}
\begin{figure}
  \centering
  \begin{minipage}{.5\textwidth}
  \includegraphics[width=\textwidth]{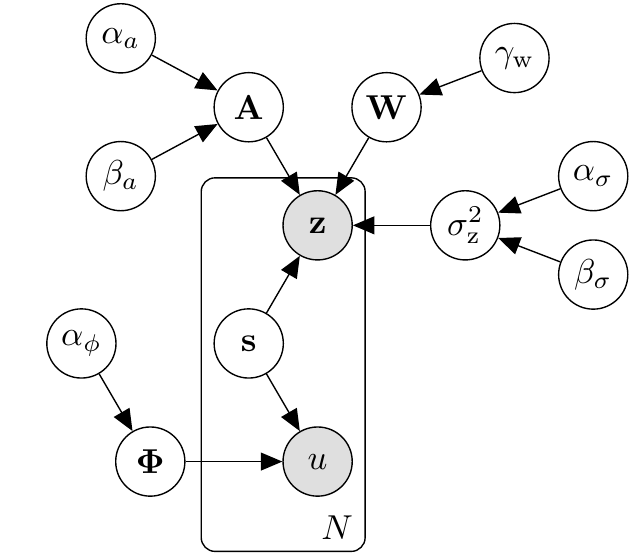} 
  \end{minipage}
  \begin{minipage}{.35\textwidth}
  \footnotesize
  \begin{tabular}{rl}
   $\zv$: & Observation
   \\ $u$: & Action
   \\ $\sv$: & substate
   \\ $\Amat$: & Activation matrix
   \\ $\Wmat$: & Weight matrix
   \\ $\Phimat$: & Policies
   \\ $\sigmaZ^2$: & Noise variance   
   \\ $\hyperAalpha, \hyperAbeta$: & Hyperparam. for $\Amat$       
   \\ $\hyperWgamma$: & Hyperparam. for $\Wmat$
   \\ $\hyperPhiAlpha{}$: & Hyperparam. for $\Phimat$
   \\ $\hypertauA, \hypertauB$: & Hyperparam. of $\sigmaZ^2$
  \end{tabular}

  \end{minipage}  
  \caption{Graphical model of the feature-based decision-making model. Only the states, $\zv_n$, and the actions, $\act_n$, with $n=1,\ldots,\numObs$, are observed. The other variables are latent and need to be inferred.}
\label{fig::model::gm}
\end{figure}

The full joint posterior distribution can be factorized as
\begin{align}
 \begin{split}
  \pdf(\Wmat, \Amat, \, & \Smat, \Phimat, \sigmaZ^2, \hyperWgamma, \hyperAalpha, \hyperAbeta, \hypertauA, \hypertauB,\hyperPhiAlpha{} \,|\, \Zmat, \actv) \propto
 \\ & \pdf(\Zmat\,|\, \Wmat, \Amat, \Smat, \sigmaZ^2) \pms(\actv\,|\,\Smat,\Phimat) \pdf(\sigmaZ^2 \,|\,\hypertauA, \hypertauB)
 \\ & \times \pdf(\Smat) \pdf(\Wmat \,|\, \hyperWgamma) \pms(\Amat \,|\, \hyperAalpha, \hyperAbeta)  \pdf(\Phimat\,|\,\hyperPhiAlpha{})
 \\ & \times \pdf(\hyperWgamma | \hyphypWgammaA, \hyphypWgammaB) \pdf(\hypertauA) \pdf(\hypertauB) \pdf(\hyperAalpha) \pdf(\hyperAbeta) \pdf(\hyperPhiAlpha{}).
 \end{split} \label{eq::model::posterior}
\end{align}
The conditional independences in this model are exploited in \secref{sec::inference}, where inference based on this model is explained. 
The structure of the posterior is illustrated as a graphical model in \figref{fig::model::gm}.

\section{Inference} 
\label{sec::inference}
We consider the problem of learning the latent variables and the prediction of optimal actions for new observations as a Bayesian inference problem. Thus, we are interested in the joint posterior distribution in \eeqref{eq::model::posterior}. Since the joint posterior is not directly tractable, we represent it by samples generated by means of Gibbs sampling. 
Therefore, in the first part of this section, we derive the conditional distributions of the variables. 
After explaining the sampling scheme in the second part, in the third part, we detail how to predict actions for new observations using this model.

For convenience, we use the bar symbol ($-$) in what follows to denote the set of conditional variables, \ie all variables except the one that shall be sampled. The set of all latent variables is denoted by 
\begin{align*}
  \Omega = \{ \Wmat, \Amat, \Smat, \Phimat, \sigmaZ^2, \hyperWgamma, \hyperAalpha, \hyperAbeta, \hypertauA, \hypertauB,\hyperPhiAlpha{} \} \in \mathbf{\Omega},
\end{align*}
where $\mathbf{\Omega}$ denotes the joint space of the latent variables.

\subsection{Conditional distributions}

\subsubsection{Sampling the noise variance}
The hyperpriors of $\hypertauA$ and $\hypertauB$ are conjugate to the prior of $\sigmaZ^2$.
Hence, the conditional $\pdf(\sigmaZ^2 \,|\, -)$ is also Inverse-Gamma distributed, 
\begin{align*}
 \pdf(\sigmaZ^2 \,|\, -)
       & \propto \invgampdf{\sigmaZ^2}{\hypertauA + \frac{\numObs D}{2} + , \hypertauB \right.
         \\ & \quad \left. + \frac{1}{2} \sumNobs \sumD \left( z_{n,d} - \sumK s_{n,k} a_{k,d} w_{k,d} \right)^2}.
\end{align*}
We use a Metropolis-Hastings algorithm with a Gamma proposal distribution to generate samples of the hyperparameters, $\hypertauA$ and $\hypertauB$.

\subsubsection{Sampling the substates}
Sampling the substates, $\Smat$, is simple thanks to the assumption of a finite space of the elements. The conditional consists of the state and action likelihoods as wells as the (conditional) prior,
\begin{align}
 \begin{split}
 \pms(s_{n,k}\,|\,-) & \propto \pdf(\zv_n \,|\, \Wmat,\Amat, \sv_n, \sigmaZ^2)
                   \pms(\act_n\,|\, \sv_n, \Phimat) \pms(  s_{n,k} \,|\, \Smat \backslash s_{n,k} ),
 \end{split}
 \label{eq::inf::s} 
\end{align}
for all $s_{n,k} \in \Sset$, where $\Smat \backslash s_{n,k}$ denotes the elements of $\Smat$ without $s_{n,k}$. As the prior for the substates follows a Beta-Binomial distribution, the conditional is simply given as 
\begin{align*}
 \pms(  s_{n,k} \,|\, \Smat \backslash s_{n,k} ) \propto \begin{cases}
			m_{s=0} + \alpha_{s=0}   \quad \text{for $s_{n,k} = 0$}
			\\ m_{s\neq0} + \alpha_{s\neq0} \quad \text{for $s_{n,k}\neq0$}
                     \end{cases},
\end{align*}
where $m_{s=0}$ and $m_{s\neq0}$ are defined in \secref{sec::model::smat}.

\subsubsection{Sampling the feature weights}
Since the likelihood is Gaussian and the weights are \iid following an Exponential distribution, the conditional of the $k$th row of the weight matrix, $\wv_k$, is a truncated Gaussian distribution,
\begin{align*}
\pdf(\wv_k\,|\,-) & \propto \pdf(\Zmat\,|\, \Wmat, \Amat, \Smat, \sigmaZ^2) \prodD \pdf(w_{k,d} \,|\, \hyperWgamma)
                  \\ & \propto \truncgausspdf{\wv_{k}}{\vs \mu_{\wv_k}, \vs \Sigma_{\wv_k}},
\end{align*}
with $\truncgausspdf{\wv_{k}}{\vs \mu_{\wv_k}, \vs \Sigma_{\wv_k}}$ denoting a truncated Gaussian, where the elements of $\wv_k$ are constraint to be positive-valued.
The mean, $\vs \mu_{\wv_k}$, and the covariance, $\vs \Sigma_{\wv_k}$, are given as
\begin{align}
    \vs \Sigma_{\wv_k}^{-1} & = \frac{1}{\sigmaZ^2} \sumNobs s_{n,k}^2 \diag{\av_k} \label{eq::inf::cov}
    \\ \begin{split} \vs \mu_{\wv_k}         & = \vs \Sigma_{\wv_k}^{-1} \Big( \frac{1}{\sigmaZ^2} \sumNobs s_{n,k} 
                        \Big( \zv_n - \sumKtwok s_{n,k'} \left(\av_{k'} \odot \wv_{k'}\right) \Big) \odot \av_{k}
                        - \onevec{D} \frac{1}{\hyperWgamma} \Big). \end{split}
\end{align}
We use the algorithm presented in \cite{Chopin2010} to sample from a truncated Gaussian distribution.
Note that we neglect the dependency on the activations in \eeqref{eq::inf::cov} since the covariance would be infinite for $a_{k,d}=0$. 
This does not affect the sampling scheme, as the corresponding weight will be ignored due to $a_{k,d}=0$ anyway.
Sampling the hyperparameter, $\hyperWgamma$, is fairly easy since the conditional of $\hyperWgamma$ is an Inverse-Gamma distribution,
\begin{align*}
 \pdf(\hyperWgamma\,|\,-) \propto \invgampdf{\hyperWgamma}{\hyphypWgammaA + \frac{KD}{2}, \hyphypWgammaB + \frac{1}{2} \sumK \sumD w_{k,d}}.
\end{align*}

\subsubsection{Sampling the feature activations}
Sampling from the \ac{IBP} consists of two steps: For each row, (i) the active columns are updated and then (ii) new features are proposed. 
In the first step, an element of the activation matrix, $a_{k,d}$, is set active with probability
\begin{align}
  \pms(a_{k,d}\,|\,-) \propto \pdf(\zv_d \,|\, \Smat \fv_d, \sigmaZ^2) \pms(a_{k,d}\,|\, \vs a_{k\backslash d} ),
  \label{eq::dm::inf::activations}
\end{align}
with $\vs a_{k\backslash d}$ denoting the $k$th row of $\Amat$ without the $d$th element and $\fv_d$ the $d$th column of $\Fmat$. 
As $\Amat$ is assumed to follow an \ac{IBP}, the conditional in \eeqref{eq::dm::inf::activations} is \cite{Ghahramani2005, Ghahramani2007}
\begin{align*}
 \pms(a_{k,d} = 1\,|\, \vs a_{k\backslash d} ) = \frac{ m_{k\backslash d}}{D + \hyperAbeta - 1},
\end{align*}
where $m_{k\backslash d}$ is the sum over $\vs a_{k\backslash d}$.

In the second step, $K^+$ new features are proposed in a Metropolis step \cite{Doshi-Velez2009, Knowles2011}.
The proposal distribution, $\proppdf{\theta^+ \,|\, \theta}$, is independent of the previous sample, $\theta$, as it consists of the priors for the substates, the feature weights, and the policies of the features,
\begin{align}
  \proppdf{\theta^+ \,|\, \theta} = \proppdf{\theta^+ } =  \pms(K^+\,|\, -) \pdf(\Wmat\,|\,\hyperWgamma) \pms(\Smat) \pdf(\Phimat\,|\,\hyperPhiAlpha{})
\end{align}
with $\theta = \{ \Wmat, \Smat, \Amat, \Phimat \}$ and $\theta^+ = \{ \Wmat^+, \Smat^+, \Amat^+, \Phimat^+ \}$, where $\Wmat^+, \Smat^+, \Amat^+$, and $\Phimat^+$ denote the proposed feature weights, activations, coefficients, and policies. 
The probability of adding $K^+$ features is given as \cite{Ghahramani2005, Ghahramani2007}
\begin{align*}
 \pms(K^+\,|\, -) \sim \poissonpdf{K^+}{\frac{\hyperAalpha \hyperAbeta}{\hyperAbeta + D - 1}}\!\!.
\end{align*}
As the proposal distribution is independent of the previous sample, the acceptance ratio, $r$, is equal to the likelihood ratio between the new and existing features \cite{Meeds2006},
\begin{align*} 
 r = \frac{ P(\Zmat \,|\, \theta^+ )}{ P( \Zmat \,|\, \theta  )}.
\end{align*}
Since the \ac{IBP} tends to mix slowly, we augment this ratio with probability $P^+$ of accepting a single new feature, resulting in a modified acceptance ratio which is derived in \cite{Knowles2011}. This increases the probability of proposing new features, leading to a faster convergence to the stationary distribution. The hyperparameters $\hyperAalpha$ and $\hyperAbeta$ are sampled as described in \cite{Knowles2011}.

\subsubsection{Sampling the policies}
Sampling from the conditionals for the policies directly is difficult and would be computationally expensive due to the mixture model (\eeqref{eq::model::mixture}).
An efficient approach is to introduce auxiliary variables, $t_n$, $n=1,\ldots,\numObs$, for each observation, indicating from which policy the observed action, $\act_n$, has been generated \cite{McLachlan2000}. 
Given the indicators, the mixture components $\phiv_k$, $k=1\ldots,K$, in \eeqref{eq::model::mixture} become conditionally independent of the mixture weights, $\sv_n$, $n=1,\ldots,\numObs$, which makes sampling the components straightforward. 
The sampling algorithm thus consists of two steps.

First, the indicators are sampled according to
\begin{align}
 \pms( t_n = k \,|\, \act_n, \Smat, \phiv ) \propto s_k \pms(\act_n \,|\, \phiv_k).
 \label{eq::inf::sample::t}
\end{align}
In order to approximate the conditional for the policies, we draw $N_t$ indicator samples from \eeqref{eq::inf::sample::t}. 
Drawing the samples is easy, since the indicators follow Categorical distributions with $t_n \in \{1, \ldots, K \}$.

Second, given the indicators, the parameters of the $k$th feature policy, $\phiv_k$, is sampled from a Dirichlet-Multinomial distribution,
\begin{align*}
  \pdf(\phiv_k\,|\,\actv,\tv) = \dirmultpdf{\phiv_k}{m_{\phi_k,1} + \hyperPhiAlpha{}, \ldots, m_{\phi_k,\numAct} + \hyperPhiAlpha{}}
\end{align*}
where $m_{\phi_k,i}$, $i=1,\ldots,\numAct$, counts the co-occurrences between the policy indicators, $\tv$, and the actions, $\act_n \in \Actset$. 

The hyperparameter, $\hyperPhiAlpha{}$, can be sampled in a Metropolis step, using a Gamma proposal distribution.

\subsection{Sampling algorithm}
\label{sec::dm::inf::sampling}
The Gibbs sampler is initialized with only one feature and the variables are sampled from their prior distributions.
After several iterations of the Gibbs sampler, samples from the target distribution are generated.

Note that there is the chance to generate new features in each iteration of the Gibbs sampler. 
Especially in scenarios with strong noise, different rows of the feature matrix may converge to similar realizations, increasing the number of features unnecessarily. 
We propose to merge features reducing the number of features, if they show a similarity larger than a prefixed threshold, $T_\text{corr}$, where we keep the activations of both features and average the policies.
The similarity is measured by means of the estimated correlation between the feature samples.
 
A \ac{MAP} estimator can be utilized, if we are interested in an estimate of the latent variables, $\Omega_{\MAP}$, containing, \ia estimates of the features, $\hat{\Fmat}_\MAP$ \cite{Rai2011}, and the policies, $\hat{\Phimat}_\MAP$.
The \ac{MAP} estimator can be approximated by choosing the sample with the highest posterior probability. 
The posterior probability is calculated according to \eeqref{eq::model::posterior}. 
For the prediction of actions for new observations, we can also use a \ac{MMSE} estimator which provides better generalization capabilities.
This is detailed in the next section.

\subsection{Prediction of actions}
\label{sec::dm::inf::prediction}
The proposed model can be used to learn the structure of the observed states as well as for the prediction of actions, given new observed states.
For the prediction of an optimal action, $\act^\star$, given a new observation, $\zv^\star$, we can evaluate the posterior predictive distribution, giving rise to a \ac{MMSE} estimator,
\begin{align} 
 \begin{split}   
 \pms(\act^\star \,|\, & \zv^\star, \Dset) 
					   = \int_{\Sset} \int_{\mathbf{\Omega}} \! \pms(\act^\star, \sv^\star \,|\, \zv^\star, \Omega) \pdf(\Omega \,|\, \zv^\star, \Dset) \, \intd\Omega \, \intd\sv^\star,
 \end{split}					  
  \label{eq::dm::post::pred}					  
\end{align}
where we exploit that $\act^\star$ is independent of the data set containing the observations, $\Dset$, given $\Omega$.
\eeqref{eq::dm::post::pred} shows that $\Omega$ depends on the new observations, $\zv^\star$. 
Thus, all variables would need to be inferred for each prediction, which would be computationally expensive. 
To remedy this issue, we assume that the observed data in $\Dset$ sufficiently represents the conditional distribution for $\Omega$, such that we can ignore the dependency and simply infer $\Omega$ based on $\Dset$, leaving only $u^\star$ and $\sv^\star$ to be inferred during prediction. 
The marginalization over $\Omega$ is approximated by Monte Carlo integration.
For this, we need to draw samples of $\sv^\star$ given the samples of $\Omega$. The samples can be generated by drawing from the conditional in \eeqref{eq::inf::s}.
Alternatively, we obtain a \ac{MAP} estimate of $u^\star$ by maximizing the posterior predictive distribution with respect to $\sv^\star$ given $\Omega_\MAP$.

Since, as justified in \secref{sec::intro::policy}, we assume a deterministic policy, the optimal action, $u^\star_\text{opt}$, is the action that maximizes $\pms(\act^\star \,|\, \zv^\star, \Dset)$. Thus, the posterior predictive distribution expresses our confidence about the actions and can be considered as the policy of the agent.

Note that, especially when we are interested in the prediction of actions, a modification on the model can help to increase the prediction accuracy dramatically.
Especially with high-dimensional observation, the observation likelihood determines the posterior, where the effect of the action likelihood nearly vanishes, leading in the worst case to a neglect of the actions.
A remedy consists in considering an action variable for each entry of the observation.
However, we assume that these actions variables are identical.
This increases the influence of the action log-likelihood by factor $\obsDim$, increasing the weight on the posterior significantly. 
The necessary modification in the inference algorithm are simple: First, for the conditional of \eeqref{eq::inf::s}, the action log-likelihood is multiplied by $\obsDim$.
Second, the same modification is applied to the joint posterior distribution in \eeqref{eq::model::posterior}.
Sampling the policies is not affected as the actions for each substate are identical.
We apply this modification on the model for the real data experiments in \secref{sec::realdata}, as the observations in the experiments are high-dimensional.

\begin{table}
 \centering
 \caption{Parameter settings for the algorithm used in the simulation experiments}
 \label{tab::res::params}
 \begin{tabular}{r|c|l}
     Parameter & Value & Meaning \\  \hline \hline
     $\hyphypSigmaAlphaA$  & 1000 & \multirow{ 4}{*}{ \!\!\!\!\! $\left. \rule{0cm}{.8cm} \right\} $ \parbox{4cm}{hyperparameters for $\sigmaZ$\\ $\hyphypSigmaAlphaA$ is chosen assuming a high \ac{SNR} scenario}} 
  \\ $\hyphypSigmaAlphaB$ & 1 &
  \\ $\hyphypSigmaBetaA$ & 1 &
  \\ $\hyphypSigmaBetaB$ & 1 &
  \\ $\hyphypAalphaA, \hyphypAalphaB$  & 1 & hyperparameters for $\hyperAalpha$
  \\ $\hyphypAbetaA$  & 1 & hyperparameter for $\hyperAbeta$
  \\ $\hyphypAbetaB$  & 10 & hyperparameter for $\hyperAbeta$
  \\ $\hyphypWgammaA, \hyphypWgammaB$ & 1 & hyperparameters for $\Wmat$
  \\ $\hyphypPhiA, \hyphypPhiB$ & 1 & hyperparameters for $\Phimat$
  \\ $\hyperSalphaA, \hyperSalphaB$ & 1 & hyperparameters for $\Smat$
  \\ $P^+$          & 0.01  & probability of accepting $K^+=1$ features
  \\ $T_\text{corr}$ & 0.9 & threshold for merging similar features
  \\ $N_\text{iter}$ & 10000 & number of iterations of the Gibbs Sampler
  \\ $L$ & 100 & size of $\Sset$
  \\ $N_t$ & 1000 & number of samples of the policy indicators
 \end{tabular}
\end{table}

\section{Experimental Results}
\label{sec::simulations}
In order to demonstrate the performance of the proposed method, we consider simulations as well as real data experiments. 
In this section, we evaluate the performance of the inference algorithm by simulating observations with different \acp{SNR} and different latent numbers of features. 
For this, we simulate the ground truth values by drawing samples from distributions which are similar to the prior distributions of the variables.
This is detailed in the following.
The true hyperparameters of the features weights are set to $\gthyphypWgammaA = \gthyphypWgammaB = 100$.
Thus, the true hyperparameter for the weights, $ \gthyperWgamma $, as well as the true weights, $ \gtWmat, $ are sampled from their prior distributions.
An element of the true feature activation matrix, $\check{\Amat}$, is activated with probability $\pms(\check{a}_{k,d}=1) =0.5$. 
The true substates, $\check{\sv}_n$, are simulated by drawing samples from a Dirichlet distribution with parameters $\frac{1}{K} \onevec{K}$, resulting in a peaked distribution.
The true noise variance, $\gtSigmaBNFL^2$, is determined by the chosen \ac{SNR}, where the signal power is estimated from $\gtXmat = \gtSmat \gtFmat$.
The parameters of the ground truth policies, $ \gtPhimat $, are chosen such that one action has high probability mass, reflecting deterministic policies, as justified in \secref{sec::intro::policy}. 
In all simulations, we consider $\numObs = 100$ observations of dimensionality $\obsDim = 30$ with $\numAct = 4$ different actions.
In order to evaluate the prediction performance of the model, we split the data set into a training and a test data set, leaving 80 observations for training and 20 for testing.
We sweep the \ac{SNR} from \unit{10}{dB} to \unit{30}{dB} with a step size of \unit{5}{dB} and vary the number of features $K$ within the set of $\{5, 7, 9, 12, 15, 18\}$. 
We set the hyperparameters of the algorithm according to \tabref{tab::res::params}.

We organize the evaluation in three parts.
First, we investigate the performance of the estimation of the features, the feature coefficients, and the reconstruction of the states. 
Second, we compare the estimated policies to the true policies and evaluate the prediction in terms of the accuracy, $ \accuracyAct $, which is the average rate of correctly predicted actions of the test data set.
These estimates are obtained utilizing a \ac{MAP} estimator.
As for the prediction of actions we can also use the \ac{MMSE} estimator, as detailed in \secref{sec::dm::inf::prediction}, we also discuss the results obtained by this estimator. 
Third, we provide results for the inferred number of features and compare it with the true value.

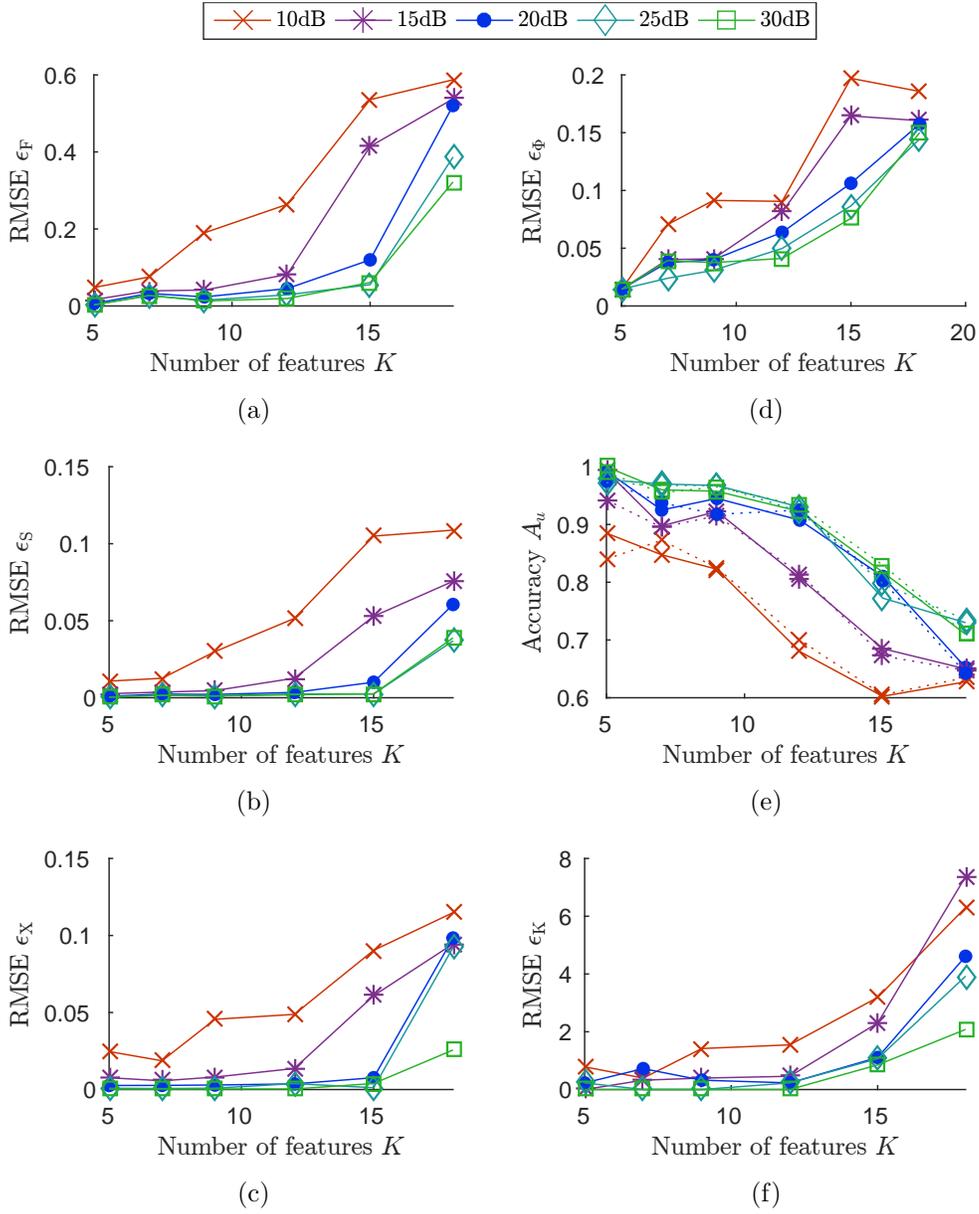
\begin{figure}
  \centering  
  \includegraphics[scale=.7]{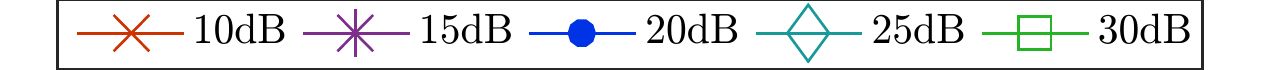}  

  \begin{tikzpicture}          
    \node[]                           (m00) {\includegraphics[width=0.45\columnwidth]{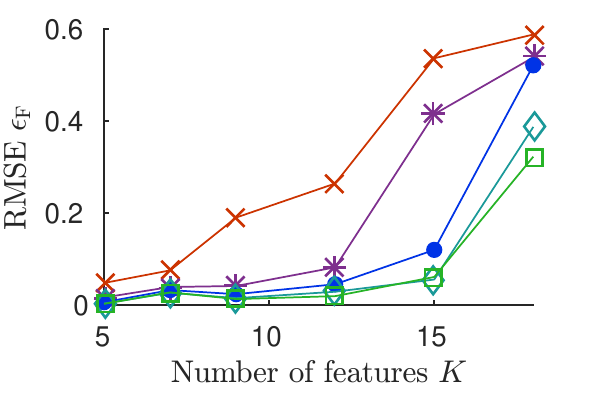}};  
    \node[below = 0cm of m00]  (d00) {(a)};
    \node[below = 0cm of d00] (m10) {\includegraphics[width=0.45\columnwidth]{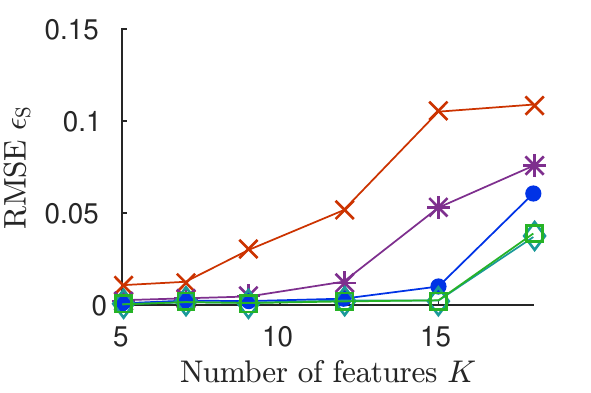}};
    \node[below = 0cm of m10]  (d10) {(b)};
    \node[below = 0cm of d10] (m20) {\includegraphics[width=0.45\columnwidth]{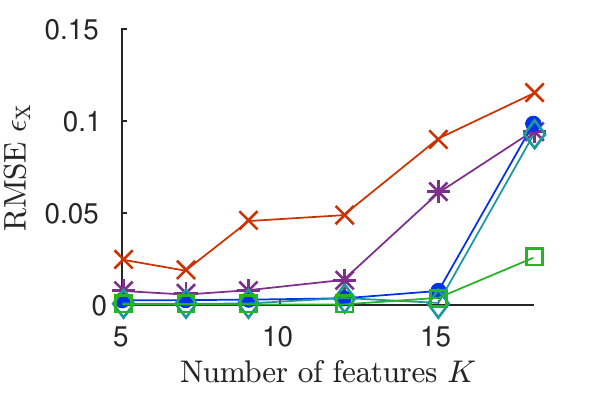}}; 		
    \node[below = 0cm of m20]  (d20) {(c)};                    	                 

    \node[right = 0cm of m00]         (m01) {\includegraphics[width=0.45\columnwidth]{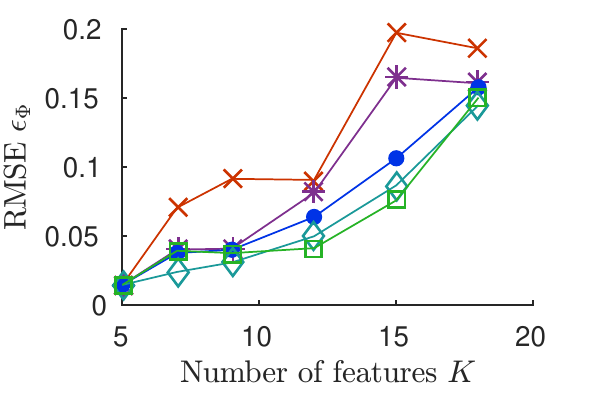}};
    \node[below = 0cm of m01]  (d01) {(d)};
    \node[below = 0cm of d01] (m11) {\includegraphics[width=0.45\columnwidth]{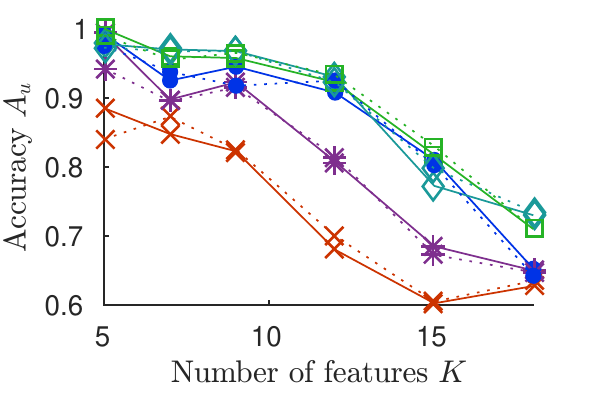}};
    \node[below = 0cm of m11]  (d11) {(e)};
    \node[below = 0cm of d11] (m21) {\includegraphics[width=0.45\columnwidth]{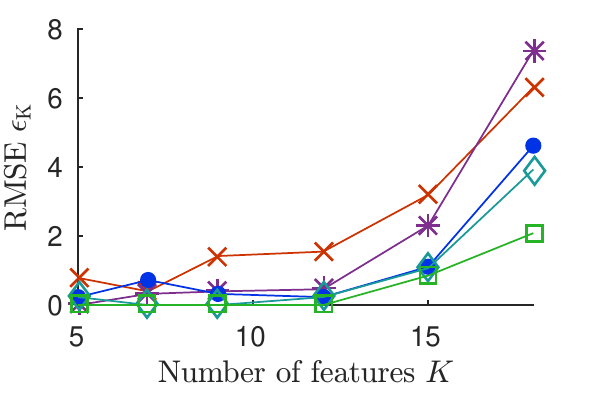}}; 		
    \node[below = 0cm of m21]  (d21) {(f)}; 
  \end{tikzpicture}  
   
  \caption{Results of the feature reconstruction. Shown are the \acp{RMSE} for (a) the features, (b) the substates, (c) the reconstructed observations, (d) the policies, (e) the accuracies of the predicted actions, and the (f) number of features. All variables are reliably inferred in case of moderate to high \acp{SNR} ($\SNR>\unit{15}{dB}$). If the \ac{SNR} drops below \unit{15}{dB}, a reliable reconstruction of more than seven features cannot be guaranteed using the simulation settings.}
  \label{fig::sim::FSZ}
\end{figure}

\subsection{Estimation of the features}
The quality of the \ac{MAP} estimates is evaluated in terms of the \ac{MSE} for the elements of the features, $\Fmat_\MAP$, the substates, $\Smat_\MAP$, and the reconstructions, $\Xmat_{\MAP} = \Smat_\MAP \Fmat_\MAP$. 
Further, we compute the \acp{RMSE} over 20 Monte Carlo runs, yielding \ac{RMSE} measures for each estimated variable, $\epsilon_\Fmat$, $\epsilon_\Smat$, and $\epsilon_{\Xmat{}}$.
The results are shown in \figref{fig::sim::FSZ}(a)--(c). 

We obtain good results with low errors especially for a small latent number of features, almost independent of the \ac{SNR}.
With increasing $K$, we observe a strong growth of the error. 
The error becomes even more significant in case of strong noise. 
It is remarkable that the error of the estimated feature values as well as the feature coefficients behave similarly.
Due to the linear relation between the features and the substates, the errors of the reconstructions also grows with the number of features.

\subsection{Action prediction}
We evaluate the correctness of the estimated policies in terms of the \ac{RMSE} of the \ac{MAD} over 20 Monte Carlo runs.
As shown in \figref{fig::sim::FSZ}(d), the error slightly grows with an increasing number of features. 
Again, the \ac{SNR} has, compared to the number of features, only little effect on the accuracy of the policies.

The accuracy of the action prediction using the \ac{MAP} and \ac{MMSE} estimators are depicted in \figref{fig::sim::FSZ}(e).
In case of few features ($K<15$) and low noise ($\SNR > \unit{20}{dB}$), we obtain highly accurate predictions with over \unit{90}{\%} accuracy. 
Only for strong noise and many latent features the accuracy drops slightly below \unit{70}{\%}.
This observation can be explained by the relation between the action and substates, which are challenging to infer in case of many features.

Since we assume in the simulation experiments that there is a fixed set of parameters that explains the observations, the \ac{MMSE} estimator yields only minor improvements over the \ac{MAP} estimator concerning the accuracy of predicted actions, as indicated by the dotted lines in \figref{fig::sim::FSZ}(e).

\subsection{Estimation of the number of features}
Assuming that the observations have been generated by a fixed number of features, we evaluate how accurately the algorithm is able to infer this number.
The results for the simulations are depicted in \figref{fig::sim::FSZ}(f), showing the \ac{RMSE} of the \ac{MAD} over the 20 Monte Carlo runs.
As can be observed, the \ac{MAP} estimator is able to infer the correct number of features reliably, especially in case of few features.
On the one hand, if the noise in the observations increases and the observations are based on many latent features, the error grows significantly.
On the other hand, if the \ac{SNR} is reasonably high, \ie $\SNR > \unit{15}{dB}$, the results deviate on average by only four from the true number in case of 18 latent features. 
The error in estimating the number of features is probably due to the fact that some simulation examples can be explained by less features than used for their generation.

\section{Real Data Experiments}
\label{sec::realdata}
\begin{figure}
 \centering
 \includegraphics[width=.4\textwidth]{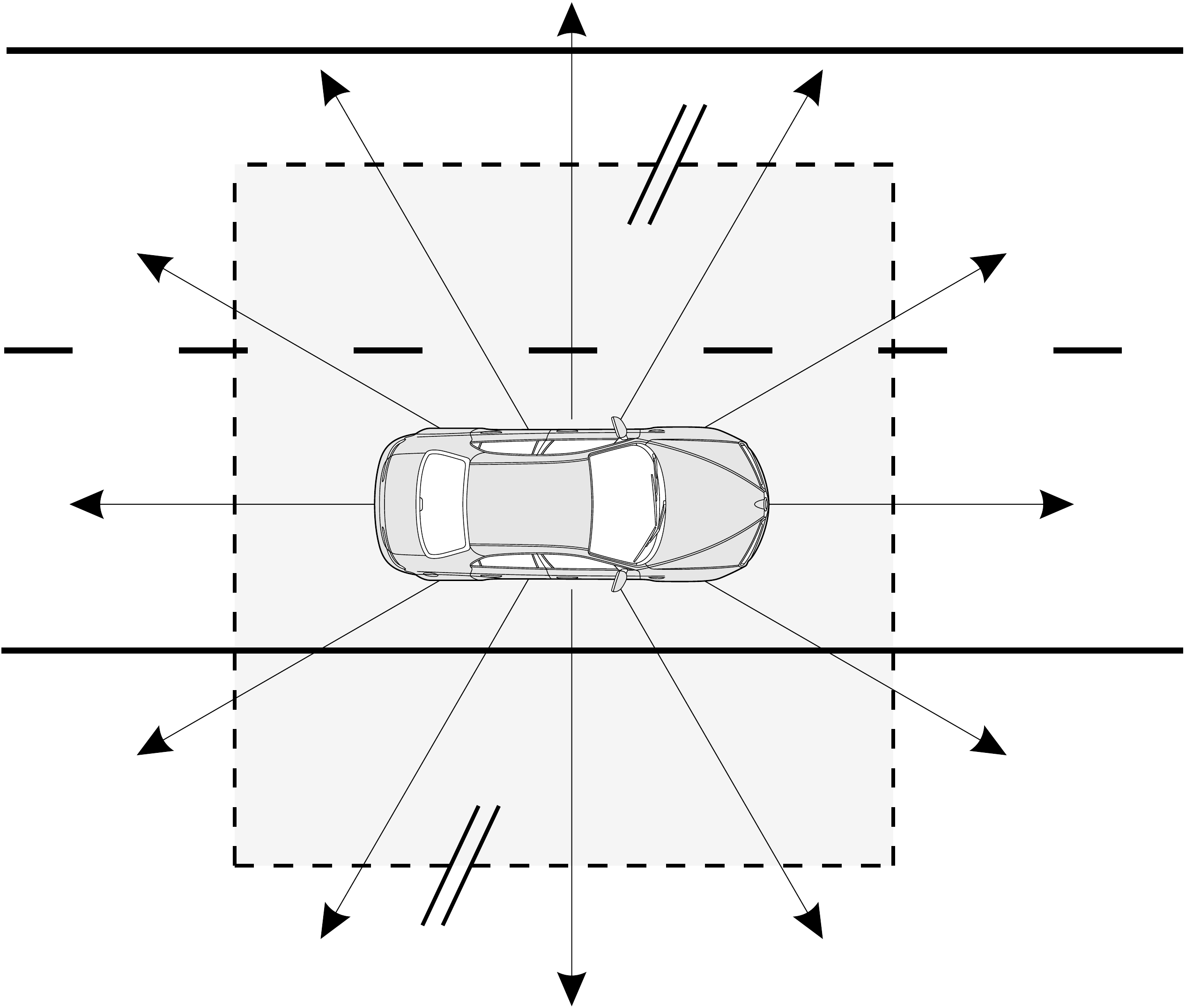}
 \caption{Illustration of the setup for the real data experiments. During pre-processing, we map the LIDAR measurement (indicated by the arrows) to an occupancy grid which is illustrated by the gray area surrounding the vehicle.}
 \label{fig::res::real::setup}
\end{figure}

We consider the problem of analyzing a driver's behavior, which is an important task for user-adaptive driver assistance systems \cite{McCall2007,Wang2016}, in order to demonstrate the performance of the proposed model in a real-world scenario.  
For this, we observe the surrounding of the vehicle and the actions taken by the driver, aiming at learning what caused the driver to make the observed decisions. 
Using the proposed model, we can also predict which action the driver is likely to take given a certain situation.
Thus, we also investigate the predictive performance of our approach by randomly creating training and test data sets.
For this, we consider real data provided by the KITTI Vision Benchmark Suite \cite{Geiger2012} containing several challenges in urban driving. 
We use the data for the tracking challenge as it contains time-sequential LIDAR measurements of different situations in public road traffic.
A schematic plot of the setup is illustrated in \figref{fig::res::real::setup}.
We consider Scene 11 and 20 of the benchmark suite, which are detailed below in \secref{sec::res::real::0011} and \secref{sec::res::real::0020}.
Before running the proposed inference algorithm, we pre-process the data as follows.
First, we apply a thresholding on the height values of the measurements such that we keep only samples above ground level, which is roughly \unit{1.5}{\meter} below the LIDAR.
Afterwards, we discretize the measurements to obtain positive-valued occupancy grids, where the values of the grid elements refer to the maximum measured height. 

As observation in the $n$th time frame, we consider the joint occupancy grid of the current and the previous frame.
This is required to implicitly include velocity information, which enables to decide, \eg if the host vehicle is faster or slower than the vehicle in front. 
Thus, the minimum speed, $v_{\min}$, between the host vehicle, $H$, and an obstacle, $O$, that can be resolved depends on the resolution, $R$, of the grid as follows,
\begin{align*}
 R \leq \abs{ d_n - d_{n-1} } = (\timeStamp{,n} -\timeStamp{,n-1}) \abs{ v_\text{H} - v_\text{O} } = \frac{1}{\sampleFreq} v_{\min},
\end{align*}
where $d_n$ denotes the distance between the host vehicle and the obstacle in the $n$th time frame. The time stamp is denoted by $\timeStamp{,n}$ and $\sampleFreq$ is the frequency at which the measurements are sampled (in the KITTI data set, $\sampleFreq = \unit{10}{\hertz}$).

We consider the environment of the vehicle in the range from $\unit{-3}{\meter}$ to $\unit{3}{\meter}$ in the lateral direction. This covers the lane of the vehicle plus parts of (if existent) neighboring lanes. In Scene 11, the longitudinal range is limited between $\unit{-10}{\meter}$ and $\unit{30}{\meter}$ and for Scene 20 between $\unit{-40}{\meter}$ and $\unit{40}{\meter}$. 
The range in longitudinal direction gives a time window of more than \unit{2}{\second} to react to the observed situation when driving at \unit{50}{\kilo\meter\per\hour}, which is the speed limit in German cities.

\begin{figure}
	\centering 
	\includegraphics[width=.45\columnwidth]{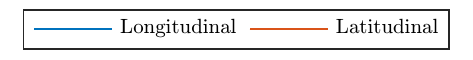}
	\begin{tikzpicture}
	\node[] (m00) {\includegraphics[width=.45\columnwidth]{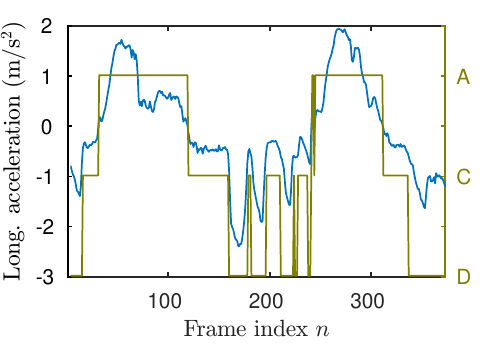}};
	\node[below =0cm of m00] (m10) {(a)};
	\node[right =0cm of m00] (m01) {\includegraphics[width=.45\columnwidth]{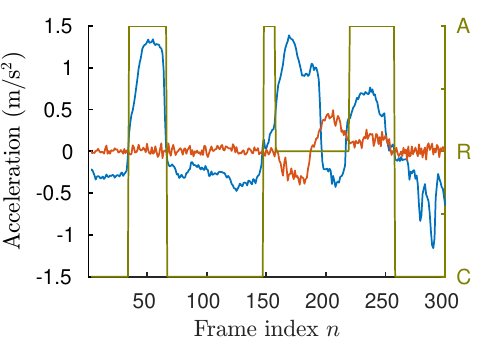}};    	
	\node[below =0cm of m01] (m11) {(b)};
	\end{tikzpicture}  
	\caption{Ground truth labels of (a) Scene 11 and (b) Scene 20. The actions \emph{acceleration} (A), \emph{lane change (right)} (R), \emph{deceleration} (D), and \emph{moving at constant speed} (C) are chosen according to the acceleration of the vehicle.}
	\label{fig::res::real::gt}
\end{figure}

For both scenes, we choose a grid size of $21 \times 65$ pixels, resulting in a spatial resolution in lateral direction of $R_\dlat \approx \unit{0.3}{\meter}$ and in longitudinal direction of $R_\dlong \approx \unit{0.6}{\meter}$ for Scene 11 (and $R_\dlong \approx \unit{1.2}{\meter}$ for Scene 20).
Thus, the minimum velocities that can be detected are $v_{\dlat, \min} > \unit{3}{\meter\per\second}$ in the lateral direction and $v_{\dlong, \min} > \unit{6}{\meter\per\second}$ in the longitudinal direction (and $v_{\dlong, \min} > \unit{12}{\meter\per\second}$ for Scene 20).

We labeled the scenes to obtain the observed actions by means of the measured acceleration of the host vehicle, obtained from the onboard inertial measurement unit. The ground truth for Scene 11 is depicted in \figref{fig::res::real::gt}(a) and for Scene 20 in \figref{fig::res::real::gt}(b).
The set of actions, $\Actset$, we consider in this experiment is comprised of \emph{acceleration} (A), \emph{deceleration} (D), \emph{lane change (right)} (R), and \emph{moving at constant speed} (C). However, due to the short duration of the sequences, usually fewer actions are observed in each scene.

We use the same settings for the hyperparameters as in \tabref{tab::res::params}. Only one of the two hyperparameter for the \ac{IBP}, $\hyphypAalphaB$, is modified to reduce the number of expected features, $\hyphypAalphaB = 10$. Further, we perform 500 iterations of the Gibbs sampler. In both data sets, after a mere 100 iterations, the sampler converges to the stationary distribution and produces samples from the target distribution.

\subsection{Scene 11 - Traffic jam}
\label{sec::res::real::0011}
Scene 11 of the KITTI data set shows an urban scenario, in which the driver follows another vehicle.
The vehicle in front cannot be overtaken due to a single-lane road.
As shown in \figref{fig::res::real::gt}(a), the driver first accelerates.
After a few seconds, the driver has to decelerate until the car stops due to a halt of the preceding car.
When the vehicle in front starts moving again, the host vehicle accelerates.

Applying the proposed algorithm to the observations results in 31 features using the \ac{MAP} estimator. 
For the analysis of the features, we only visualize three of the most relevant features, which are selected according to the confidence in the corresponding feature policy.

Considering the current and the previous frames as inputs, we obtain feature estimates of consecutive frames.
Observing the differences between the frames can be difficult as the features are likely to be highly similar due to the low temporal difference.
Therefore, we show the feature of the current time frame and the difference of the features. 
A difference plot can be interpreted as the temporal gradient of two consecutive frames. 
Thus, from the depicted patterns we can draw conclusions about the relative speed of the obstacles. 
For instance, if the difference plot shows negative values left of positive values (red-blue pattern), the obstacle is faster than the host vehicle. 
In contrast, a blue-red pattern indicates a slower vehicle. 
The width of the bars reflect the magnitude of the relative velocities.

\begin{figure}
	\centering 	
	\begin{tikzpicture}          
	\node[                 ] (m00) {\includegraphics[width=0.3\columnwidth]{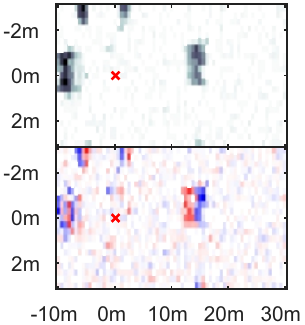}};
	\node[right =0cm of m00] (m01) {\includegraphics[width=0.3\columnwidth]{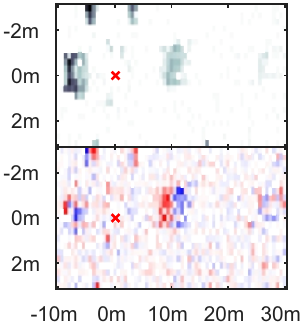}};
	\node[right =0cm of m01] (m02) {\includegraphics[width=0.3\columnwidth]{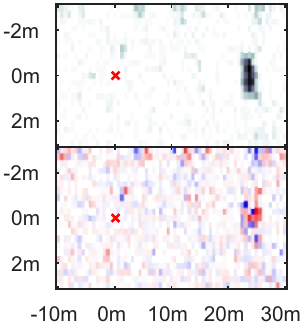}};
	\node[above =0cm of m00] {\small A1};
	\node[above =0cm of m01] {\small A2};
	\node[above =0cm of m02] {\small A3};
	
	\node[below =0.5cm of m00] (d10) {\small D1};
	\node[below =0.5cm of m01] (d11) {\small D2};
	\node[below =0.5cm of m02] (d12) {\small D3};	
	\node[below =0cm of d10] (m10) {\includegraphics[width=0.3\columnwidth]{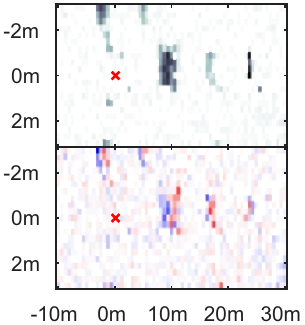}};
	\node[right =0cm of m10] (m11) {\includegraphics[width=0.3\columnwidth]{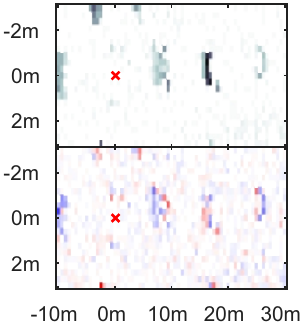}};
	\node[right =0cm of m11] (m12) {\includegraphics[width=0.3\columnwidth]{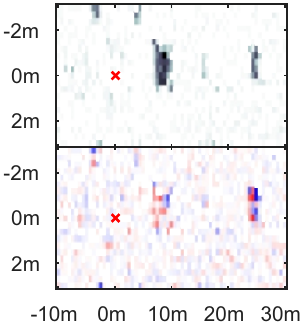}};
	
	\node[below =0.5cm of m10] (d20) {\small C1};
	\node[below =0.5cm of m11] (d21) {\small C2};
	\node[below =0.5cm of m12] (d22) {\small C3};	
	\node[below =0cm of d20] (m20) {\includegraphics[width=0.3\columnwidth]{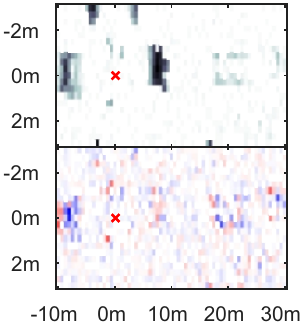}};
	\node[right =0cm of m20] (m21) {\includegraphics[width=0.3\columnwidth]{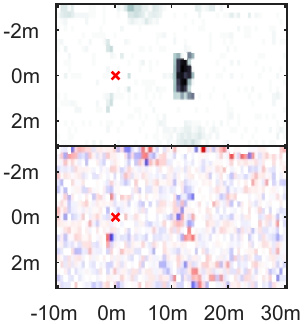}};
	\node[right =0cm of m21] (m22) {\includegraphics[width=0.3\columnwidth]{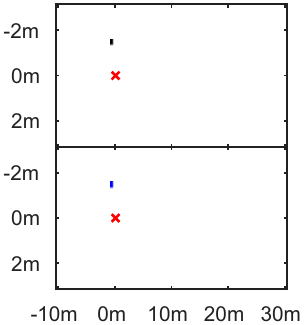}};
	\end{tikzpicture}  
	
	\caption{Features estimated for KITTI Scene 11. The host vehicle is indicated by the red cross. Depicted are the features (top) and the difference plots (bottom), where blue indicates positive and red negative values. First row: action A (acceleration). Features A1 and A2 show an obstacle in front of the host vehicle, which is moving significantly faster as indicated by the difference plots. A3 shows an obstacle that slows down. As it is sufficiently far away, the driver still accelerates. Second row: action D (deceleration). Features D1 and D2  show vehicles in front of the host car, which are moving at a slower speed. D3 shows that the preceding car accelerates. As the host vehicle is close to the obstacle, the driver decelerates. Third row: action C (moving at constant speed).  Features C1 and C2 show preceding vehicles. As the difference plots do not indicate significant speed differences, the preceding and the host car travel at a similar speed, such that the driver does not need to adapt the speed. C3 shows an empty road without any obstacles.}
	\label{fig::res::0011::feats}
\end{figure}
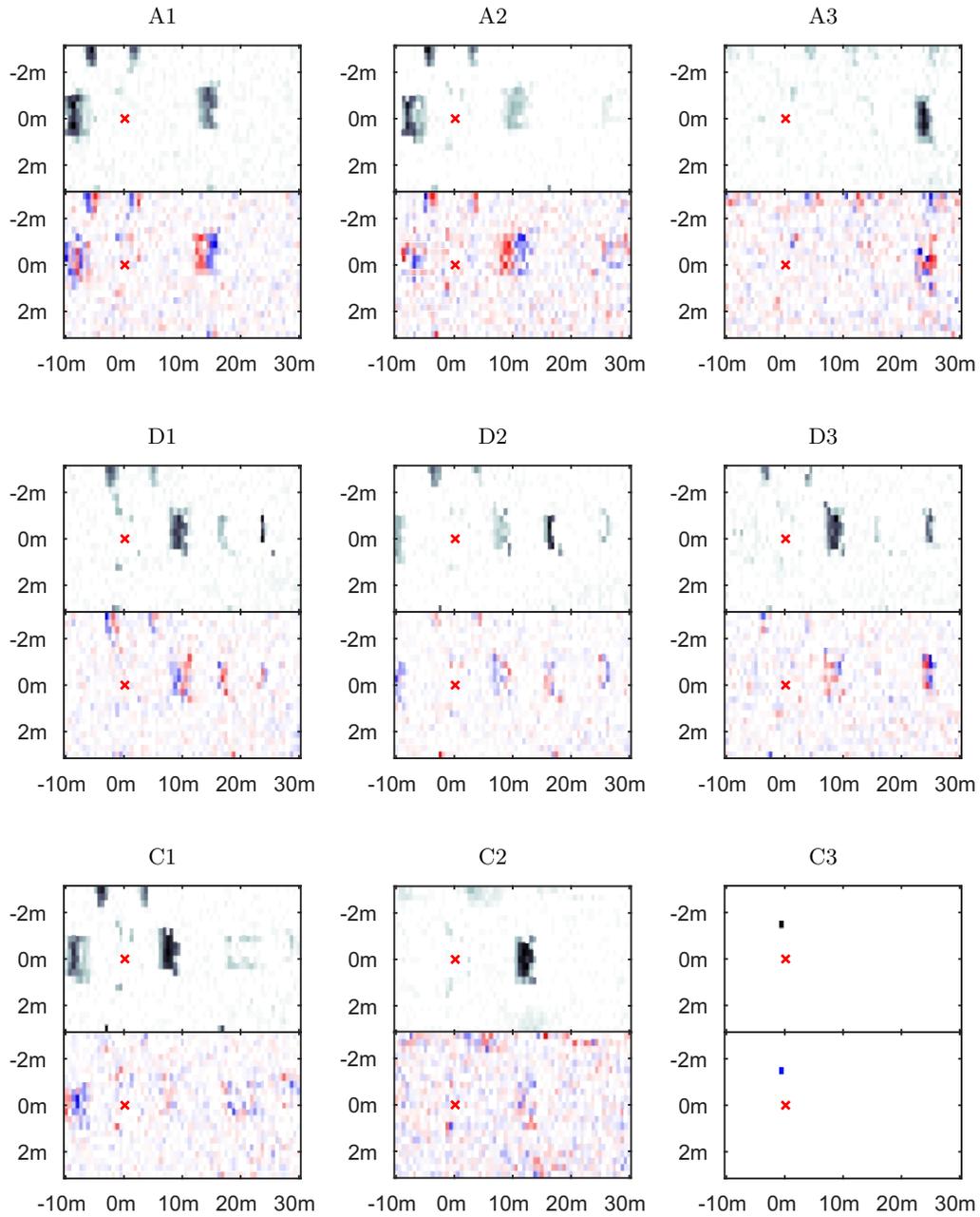

The features are illustrated in \figref{fig::res::0011::feats}.
In the figures, the $x$-axis corresponds to the longitudinal and the $y$-axis to the lateral direction.
Shown is the top-view on the host vehicle, which is indicated by the red cross.
The intensity reflects the measured heights at each pixel.
As explained, the difference features are obtained by subtracting the features of the previous from the current time frame. 
Features A1 and A2 clearly show that the driver accelerates as long as the preceding car is significantly faster. A3, however, indicates acceleration though the preceding car decelerates. Since the car seems to be sufficiently far away, the driver has not yet decided to reduce the speed.
Features D1 and D2 explain the deceleration of the driver with a slower car in front. In contrast, D3 shows that the preceding car accelerates. However, due to the low distance between both cars, the driver decides to decelerate.
Features C1 and C2 show a vehicle in front of the driver's car. The difference images reveal only little differences between the velocities of both vehicles, such that the driver maintains the current velocity. C3 represents an empty road, where, again, the driver does not need to adapt the speed of his car.

In order to evaluate the prediction performance, as in the simulations, 20\% of the observations are used for testing while the rest is used for inferring the structure. 
Using the \ac{MAP} estimator results in an accuracy of $74.32$ \%. 
The confusion matrix in \tabref{tab::res::real::conf::0011} shows that, most notably, in some cases moving at constant speed and acceleration are confused. 
Further, deceleration is in few cases misclassified as acceleration.

Quantifying the results of the state reconstruction is difficult, as a ground truth is not available. 
To provide at least an intuition about the quality of the reconstructions, we compute the \ac{MSE} between the reconstructed states, based on the estimated substates and features, and the observations.
This results in a \ac{MSE} of approx. $0.0329$. 
As the values of the observations are in a similar range as in the simulations, this result indicates low errors, resembling the results of the simulations.
Still, the number has to be taken with care, as we compare the denoised reconstruction with noisy observation. 

\begin{table}
 \centering
 \caption{Confusion matrix for the hold-out data set of Scene 11. The overall accuracy is $74.32\%$, where 31 features have been inferred.}
 \begin{tabular}{r||ccc}
       & \multicolumn{3}{c}{Prediction}
   \\ Ground truth  & Const. speed  & Deceleration & Acceleration
   \\ \hline \hline
      Const. speed & \bf 12 & 2 & 4
   \\ Deceleration & 1  & \bf 15 & 6
   \\ Acceleration & 5  & 1 & \bf 28
 \end{tabular}
 \label{tab::res::real::conf::0011}
\end{table}

\subsection{Scene 20 - Lane change}
\label{sec::res::real::0020}
Scene 20 shows a lane change maneuver on a two-lane road.
As can be observed from the acceleration signals depicted in \figref{fig::res::real::gt}(b), the driver accelerates three times, depending on the current traffic situation.
The lane change takes place from time frame 158 to 220.

Applying the proposed algorithm reveals 17 features. Using these features to predict the actions of the test data set yields an accuracy of $73.33\%$. The confusion matrix in \tabref{tab::res::real::conf::0020} shows that lane changes are reliably predicted.
Acceleration maneuvers are in some cases misinterpreted as moving at constant speed or as a lane change.
A third of the moving at constant speed observations are misclassified as lane changes.
Comparing the reconstructed states with the noisy observation yields a \ac{MSE} of $0.0027$.

\begin{figure}
	\centering 	
	\begin{tikzpicture}          
	  \node[               ] (m00) {\includegraphics[width=.3\columnwidth,trim={0cm, 0.4cm, 0.2cm, 0.6cm},clip]{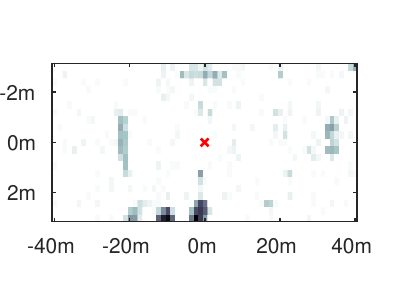}};
	  \node[below =0 of m00] (d00) {\small (a)};
	  \node[right =0 of m00] (m01) {\includegraphics[width=.3\columnwidth,trim={0cm, 0.4cm, 0.2cm, 0.6cm},clip]{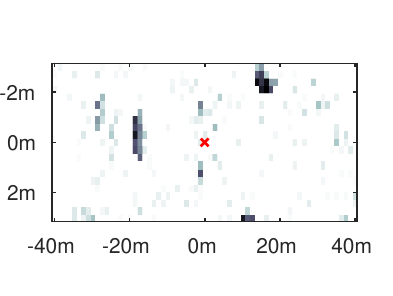}};
	  \node[below =0 of m01] (d01) {\small (b)};
	  \node[right =0 of m01] (m02) {\includegraphics[width=.3\columnwidth,trim={0cm, 0.4cm, 0.2cm, 0.6cm},clip]{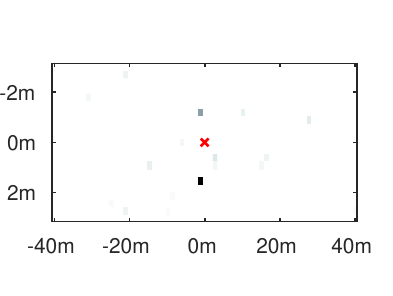}};
	  \node[below =0 of m02] (d02) {\small (c)};
	\end{tikzpicture}  

	\caption{Features estimated from KITTI Scene 20. Shown are 3 out of 17 inferred features and the host vehicle (red cross). The first feature (a) indicates a lane change (right). The scene is rotated, showing two vehicles behind the host vehicle in the lower left corner. The second feature (b) represents moving at constant speed, most likely due to high traffic density (vehicle behind the host and on the left lane). The third feature (c) indicates acceleration and shows a free lane. }
	\label{fig::res::0020::feats}
\end{figure}
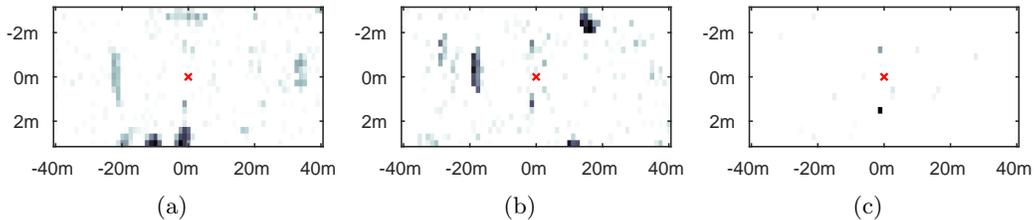

For illustration of the inferred features, \figref{fig::res::0020::feats} shows three out of the 17 features, each indicating a different action. The first feature (a) shows a lane change (right). 
As the driver is already performing the maneuver, the scene is rotated. The dark areas in the lower left corner represent vehicles behind the host car. 
The second feature (b) contains a vehicle behind the driver's car and another vehicle on the left lane. Due to the high traffic density, the driver does not accelerate.
In contrast, the third feature (c) shows a completely empty road, motivating the driver to accelerate the vehicle.

\begin{table}
 \centering
 \caption{Confusion matrix for the hold-out data set of Scene 20. The overall accuracy is $73.33\%$, where 17 features have been inferred. }
 \begin{tabular}{r||ccc}
       & \multicolumn{3}{c}{Prediction}
   \\ Ground truth  & Acceleration  & Lane change (right) & Const. speed
   \\ \hline \hline
      Acceleration        & \bf 3 & 3 & 7
   \\ Lane change (right) & 0  & \bf 29 & 0
   \\ Const. speed              & 0  & 6 & \bf 12
 \end{tabular}
 \label{tab::res::real::conf::0020}
\end{table}

\section{Discussion}
\label{sec::discussion}
As shown in the simulation experiments, the algorithm based on the proposed model is able to reliably infer the number of features, the features and the policies. 
In case of strong noise, the algorithm finds several, almost equally probable explanations for the observations, resulting in variations of the MAP estimate. 
Especially when the number of features is high compared to the dimensionality of the observations, inferring the correct number of features can be challenging.

The real data experiments show that the model is able to provide deeper insights into the observations which may yield new conclusions about the observed behavior. 
As explained, in high-dimensional observations, the observation likelihood is likely to dominate the posterior leading to only little influence of the action likelihood and, hence, poor prediction performance. 
As proposed in \secref{sec::dm::inf::prediction}, reweighting the action likelihood, assuming an action variable for each entry of the observation, yields a significant performance increase, while the observed states can still be reliably reconstructed.

An advantage of the proposed generative model is that it can be modified and extended easily. 
This is helpful especially for a different assumption on the feature weights. 
For example, if real-valued features are expected, the corresponding prior can be changed to a Gaussian distribution. 
Of course, inference has to be adapted accordingly.
Further, one can easily extend the proposed model to a semi-supervised learning approach, in which we add variables for states where the actions are not observed. 
Thus, the state representations can be learned from even more data resulting in more accurate estimates.

\section{Conclusion}
\label{sec::conclusion}
We have presented a feature-based framework for learning from demonstrations that allows us to reason about the observed behavior and to predict actions for new states.
A key assumption in the proposed model is that the observations are composed of latent features.
Each feature imposes its own policy and contributes to the decision of the agent.
To learn the structure of the behavior and to predict actions, we have considered a Bayesian nonparametric approach based on the \acl{IBP}, which allows to infer the number of features and the features itself from the observed data. 
By means of this model, we are able to obtain a deeper understanding of the observed behavior as the features and their policies allow to reason about the observed decisions. 
The simulations show that the developed algorithm performs well. 
Only in scenarios with strong noise and many latent features, inference becomes challenging.
Further, we have considered the task of learning a driver's behavior. 
For this, we have applied our algorithm to real data obtained from the KITTI benchmark suite. 
The results reveal under which conditions the driver takes the observed actions. 
Further, prediction on a hold-out data set demonstrates that actions can be predicted reliably.

\bibliographystyle{plain}
\bibliography{pr_bib}

\end{document}